\documentclass[runningheads]{llncs}
\usepackage{enumitem}
\usepackage{todonotes}

\usepackage{comment}
\usepackage{graphicx}
\usepackage{subfig}
\usepackage{wrapfig}
\usepackage{amsmath}
\usepackage{array}
\newcolumntype{L}[1]{>{\raggedright\let\newline\\\arraybackslash\hspace{0pt}}m{#1}}
\newcolumntype{C}[1]{>{\centering\let\newline\\\arraybackslash\hspace{0pt}}m{#1}}

\newcommand\negspace{-15pt}
\begin{document}
%
\title{Causal Discovery of Dynamic Models for Predicting Human Spatial Interactions
\thanks{This work has received funding from the European Union's Horizon 2020 research and innovation programme under grant agreement No 101017274 (DARKO).}}

%
\author{
Luca Castri\inst{1} \and
Sariah Mghames\inst{1} \and
Marc Hanheide\inst{1} \and
Nicola Bellotto\inst{1,2}
}
\authorrunning{L. Castri et al.}
%
\institute{University of Lincoln, UK, 
\email{\{lcastri,smghames,mhanheide\}@lincoln.ac.uk}\\\and University of Padua, Italy,
\email{nbellotto@dei.unipd.it}}
\maketitle              

\begin{abstract}
Exploiting robots for activities in human-shared environments, whether warehouses, shopping centres or hospitals, calls for such robots to understand the underlying physical interactions between nearby agents and objects. In particular, modelling cause-and-effect relations between the latter can help to predict unobserved human behaviours and anticipate the outcome of specific robot interventions. 
In this paper, we propose an application of causal discovery methods to model human-robot spatial interactions, trying to understand human behaviours from real-world sensor data in two possible scenarios: humans interacting with the environment, and humans interacting with obstacles.
New methods and practical solutions are discussed to exploit, for the first time, a state-of-the-art causal discovery algorithm in some challenging human environments, with potential application in many service robotics scenarios. 
To demonstrate the utility of the causal models obtained from real-world datasets, we present a comparison between causal and non-causal prediction approaches.
Our results show that the causal model correctly captures the underlying interactions of the considered scenarios and improves its prediction accuracy.
\end{abstract}

\section{Introduction} \label{sec:intro}
The increased use of robots in numerous sectors, such as industrial, agriculture and healthcare, represents a turning point for their progress and growth. However, it requires also new approaches to study and design effective human-robot interactions. 
A robot, sharing the working area with humans, must accomplish its task taking into account that its actions may lead to unpredicted responses by the individuals around it. Knowing the cause-effect relationships in the environment will allow the robot to reason on its own actions, which is a crucial step towards effective human-robot interactions and collaborations.

\begin{figure}[t]\centering
\includegraphics[trim={0cm 3.6cm 0cm 3cm}, clip, width=\columnwidth]{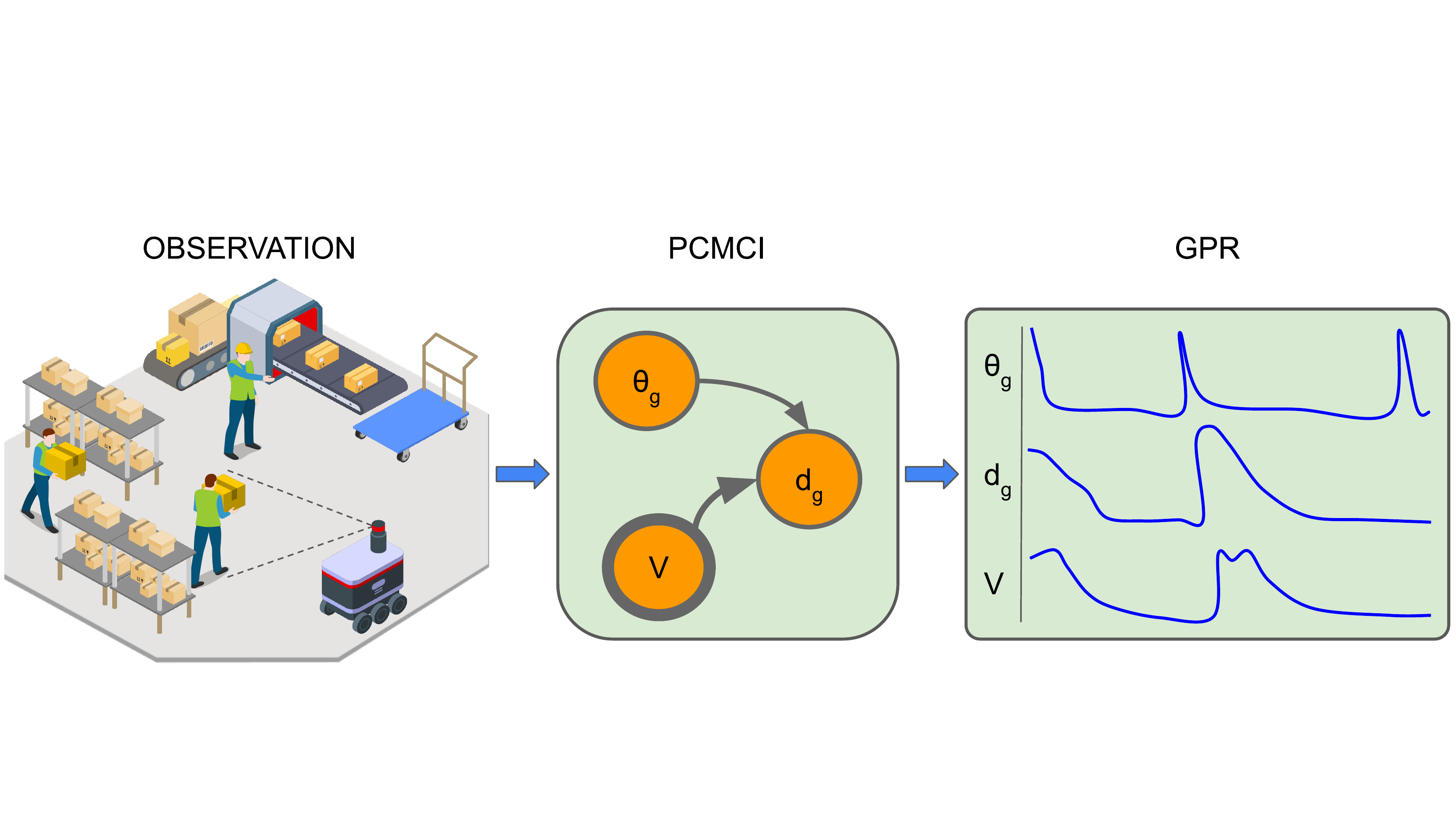}
\vspace{-18pt}
\caption{\small{Causal prediction approach: a robot reconstructing a causal model from observation of human behaviours in a warehouse environment. The causal model is then used for human spatial behaviour prediction.}}
\vspace{\negspace}
\label{fig:intro}
\end{figure}

Causal inference, which includes causal discovery and reasoning, appears in the literature of different fields, including robotics~\cite{hellstrom_relevance_2021,brawer_causal_2021,cao_reasoning_2021,Katz2018,Angelov2019}.
However, most of the work on human-human and human-robot spatial interactions, i.e. the 2D relative motion of two interacting agents~\cite{bellotto2013qualitative,Dondrup2014,hanheide2012analysis} did not previously exploit any formal causal analysis. 
Most mobile robots do not know how humans will behave and react as a consequence of their proximity and actions. Knowing the causal model of the interactions between different agents could help predict human motion behaviours, and consequently assist a robot planner in choosing the most effectively navigation strategy.
For instance, a robot in a warehouse environment (e.g. see Fig. 1), passing very close to a human, needs to know how the human would react to this situation in order to choose the most appropriate behaviour (e.g. "I can continue since the human will remain still" or "it is better to stay still and wait for the human to go away"). More generally, discovering the causal model will enable the robot to assess future interventions (e.g. “what happens if I go this way?”) and counterfactual situations (e.g. “what would have happened if I remained still instead of moving?”).

In this paper, we demonstrate that by using a suitable causal discovery algorithm, a robot can estimate the causal model of the nearby human motion behaviours by observing their trajectories and spatial interactions. In particular, our main contributions are the following:
\vspace{-5pt}
\begin{itemize}
    \item first application of a causal discovery method to real-world sensor data for modeling human and robot motion behaviours, with a focus on 2D spatial interactions;
    \item new causal models to represent and predict humans-goal, human-human and human-robot spatial interactions, in single and multi-agent scenarios;
    \item experimental evaluation of the causal models on two challenging datasets to predict spatial interactions in human environments.
\end{itemize}
\vspace{-5pt}
The paper is structured as follows: basic concepts about causal discovery and its applications are presented in Section \ref{sec:related}; Section \ref{sec:appr} explains the details of our approach; the application of our approach in real-world scenarios is described in Section \ref{sec:exp}, including also comparison between the experimental results; finally, we conclude this paper in Section \ref{sec:conc} discussing achievements and future applications.

\section{Related work} \label{sec:related}

Modeling human motion behaviours and spatial interactions is an important research area. In~\cite{Liu2022}, the authors introduce a high level causal formalism of motion forecasting, including human interactions, based on a dynamic process with different types of latent variables to take into account also unobserved and spurious features. 
Another approach to model spatial interactions in social robotics is by using a qualitative trajectory calculus (QTC) to explicitly account for motion relations between human-human and human-robot pairs, such as relative distance, direction, and velocity~\cite{bellotto2013qualitative,Dondrup2014,hanheide2012analysis}. In this case though, the causal links between spatial relations were never taken into account. 
Our current work is inspired by the same QTC relations, extended to include other factors (e.g. collisions) and represented in quantitative rather than qualitative terms.

Among possible causal representations, Structural Causal Models~(SCMs) and Directed Acyclic Graphs~(DAGs) are the most popular ones~\cite{alma991011292629705181}. The latter consist of nodes and oriented edges to represent, respectively, variables and causal dependencies between them (see Fig.~\ref{fig:intro}). 
Several methods have been recently developed to derive causal models from observational data, a process termed {\em causal discovery}. They can be categorised into two main classes~\cite{glymour_review_2019}: constraint-based methods, such as Peter \& Clark~(PC) algorithm and Fast Causal Inference~(FCI), and score-based methods, such as Greedy Equivalence Search~(GES). Recently, reinforcement learning-based methods have also been used to discover causal models~\cite{Zhu2019,Gasse2021}. 
However, many of these algorithms work only with static data (i.e. no temporal information), which is a limitation in many robotics applications.
In fact, methods for time-dependent causal discovery are necessary to deal with time-series of sensor data.
To this end, a variation of the PC algorithm, called PCMCI~\cite{runge_causal_2018}, was proposed to efficiently reconstruct causal graphs from high-dimensional time-series datasets, which is based on a false positive rate optimisation and a momentary conditional independence~(MCI) test.

PCMCI applications can be found in climate and healthcare sectors~\cite{runge_detecting_2019,saetia_constructing_2021}.
Other key concepts of causal inference extended to the machine learning domain can also be found in~\cite{scholkopf_toward_2021,Seitzer2021}.
In robotics, recent works include a method to build and learn a SCM through a mix of observation and self-supervised trials for tool affordance with a humanoid robot~\cite{brawer_causal_2021}.
Another application includes the use of PCMCI to derive the causal model of an underwater robot trying to reach a target position~\cite{cao_reasoning_2021}. 
Other causal approaches can also be found in the area of robot imitation learning~\cite{Katz2018,Angelov2019}. 

To our knowledge though, none of the above applications have explored causal discovery to understand human-human and human-robot spatial interactions. Our goal indeed is for the robot to recover cause-and-effect in human motion behaviours when they collaborate and share the same environment. To this end, we will derive some useful causal models of spatial interactions in particular scenarios and use them to predict the occurrence of future ones.

\section{Causal discovery from observational data} \label{sec:appr}
Our approach is based on the observation of human spatial behaviours to recover the underlying SCM. This causal analysis is performed by using the PCMCI causal discovery algorithm \cite{runge_causal_2018}. First, we identify some important factors (i.e. variables) affecting human motion in the considered scenarios, and from that we reconstruct the most likely causal links from real sensor data. Finally, we use the discovered causal models to forecast the latter with a state-of-the-art Gaussian Process Regression (GPR) technique~\cite{li2020data}, showing that the causality-based GPR improves the accuracy of the human (interaction) prediction compared to a non-causal version.
Two different scenarios have been modelled and analysed.

\begin{figure}[b]
\centering 
\subfloat{
    \label{fig:exp_thor}
    \includegraphics[width=0.32\columnwidth]{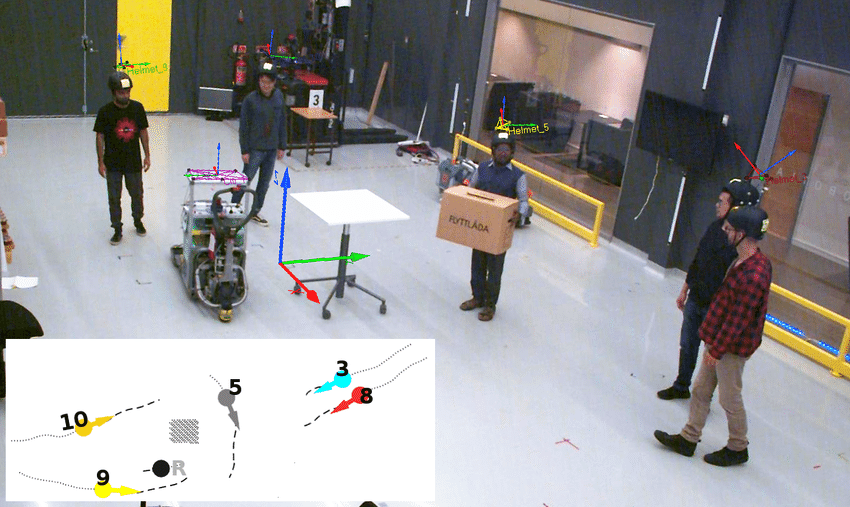}}
\subfloat{
    \label{fig:approach_hg_scenario}    \includegraphics[width=0.32\columnwidth]{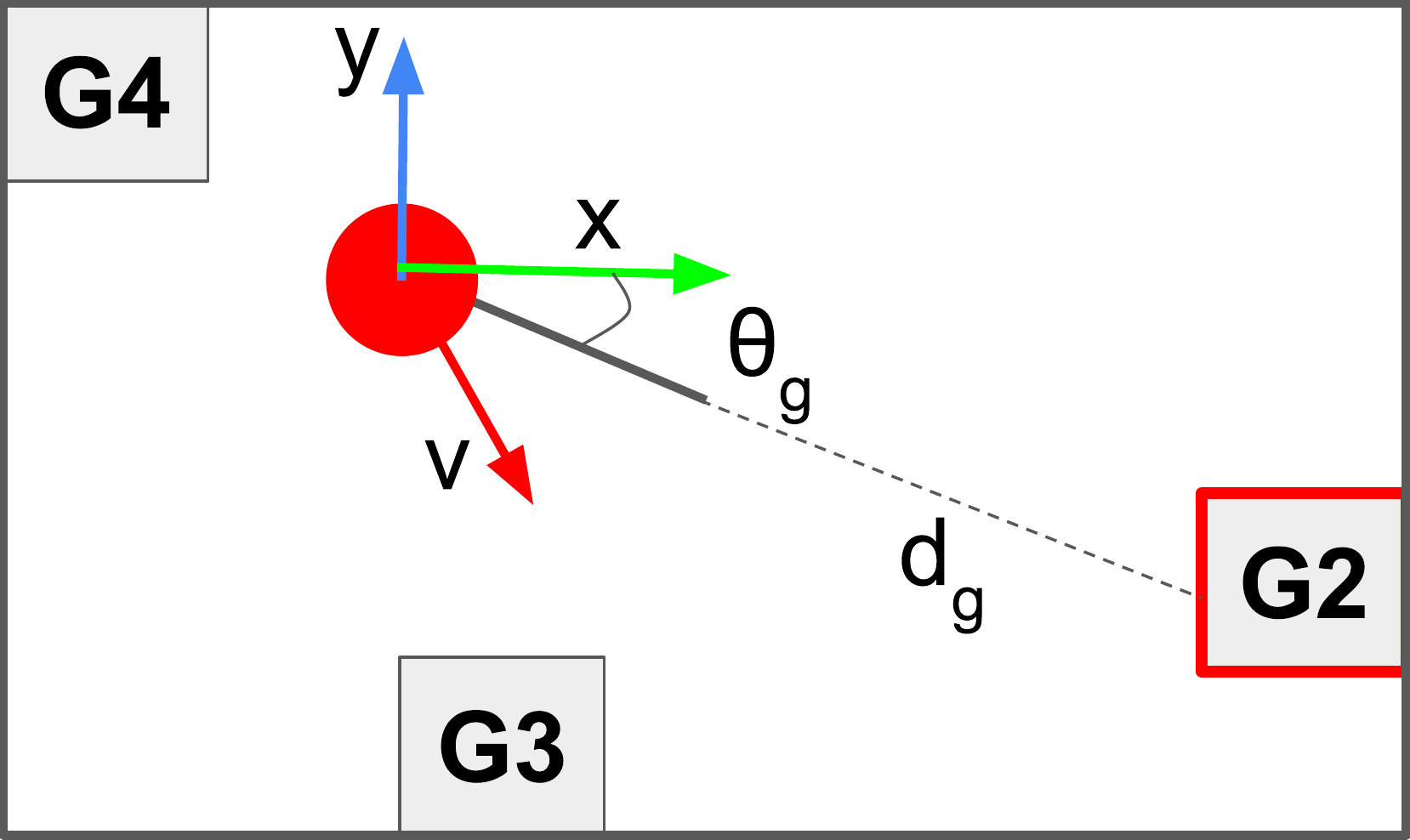}}
\subfloat{
    \label{fig:approach_hmo_scenario}
    \includegraphics[width=0.32\columnwidth]{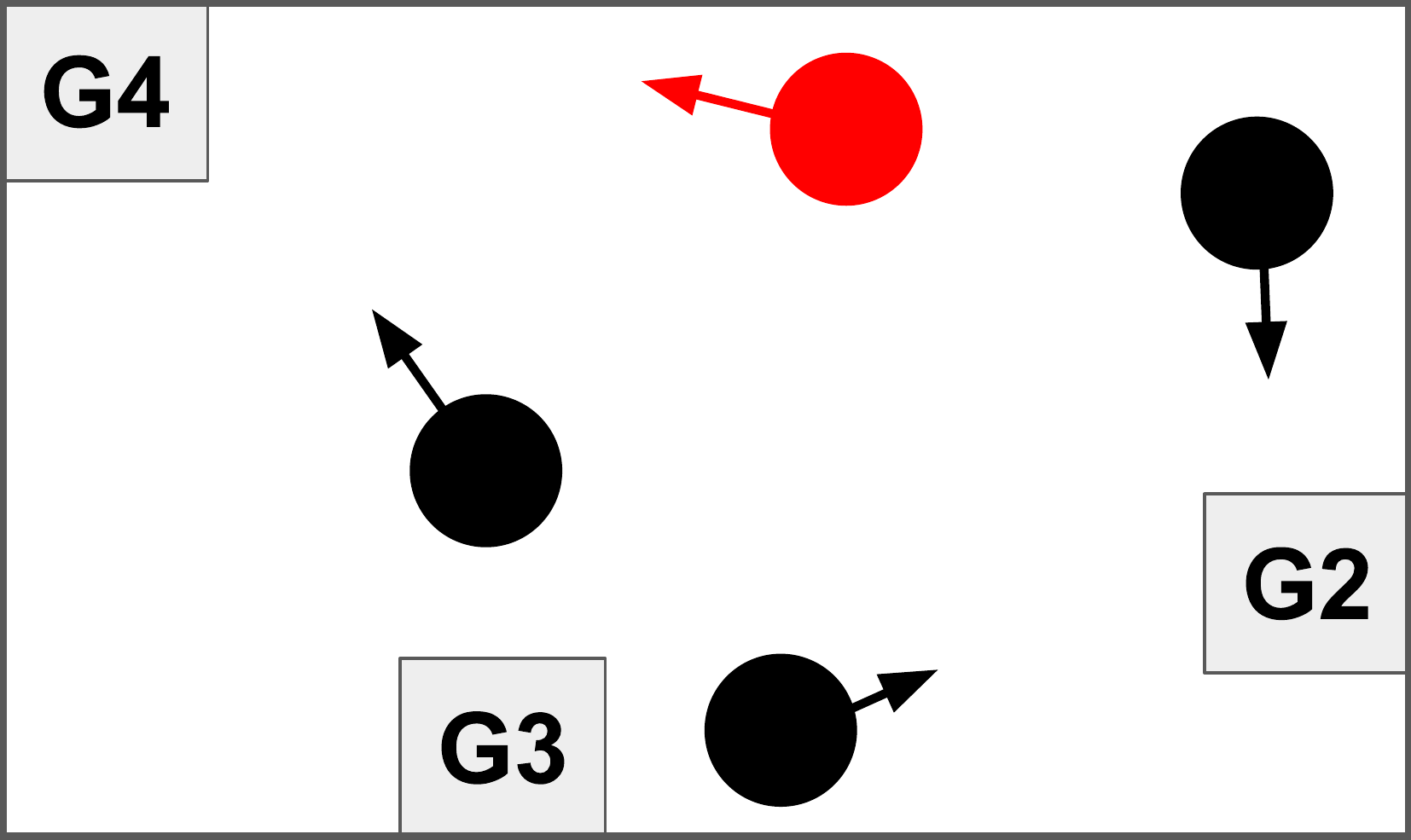}}
\vspace{-7pt}
\caption{\small{Image from TH{\"O}R dataset \cite{thorDataset2019} (left). Representation of the two analysed scenarios: (centre) the human-goal scenario, (right) the human-moving obstacles scenario. The agents consist of a circle and an arrow specifying, respectively, the current position and orientation. The selected agent is red, while, the obstacles are black.}}
\vspace{\negspace}
\label{fig:approach_scenarios}
\end{figure}

\subsection{Human-goal scenario} \label{subsec: Human-goal scenario}
Our first scenario includes interactions between human and (static) goals in a warehouse-like environment, illustrated in Fig.~\ref{fig:approach_scenarios}~(centre), where the agent walks among different positions (grey squares) to move some boxes or grab/use some tools. The grey line connecting agent and goal specifies the angle~$\theta_g$ between the two. Upon expert judgment, the following features were deemed essential to explain the human motion behaviour: \textit{(i)}~angle agent-goal $\theta_g$; \textit{(ii)}~euclidean distance agent-goal $d_g$; \textit{(iii)}~agent velocity $v$.
The angle $\theta_g$ represents the human intention to reach a desired position (the person will first point towards the desired target before reaching it); then the person walks towards the goal, reducing the distance from it, at first by increasing the walking speed and finally decreasing, when close to the destination. Soon after the human has reached the goal, $\theta_g$ changes to the next one, and the process restarts. 
What we expect from this scenario are therefore the following causal relations:
\vspace{-5pt}
\begin{enumerate}[label=(\alph*)]
    \item $\theta_g$ depends on the distance, when the latter decreases to zero then $\theta_g$ changes;
    \item $d_g$ is inversely related to $v$ and depends on $\theta_g$;
    \item $v$ is a direct function of the distance $d_g$. 
\end{enumerate}
\vspace{-5pt}
\subsection{Human-moving obstacles scenario} \label{subsec: Human-moving obs scenario}
The second scenario involves multiple agents. It reproduces the interaction between a selected human and nearby dynamic obstacles (e.g. other humans, mobile robot), as shown in Fig.~\ref{fig:approach_scenarios}~(right). In this case, we take into account human reactions to possible collisions with obstacles, modelled by a \textit{risk} factor.
Consequently, the relevant features in this scenario are \textit{(i)}~euclidean distance~$d_g$ of the selected agent-goal, \textit{(ii)}~agent's velocity $v$, and \textit{(iii)}~\textit{risk} value.
The agent moves between goals in the environment, so the cause-effect relation between distance and velocity will be similar to the previous scenario. The main difference in this case is that, instead of reaching the goal without problems, the agent needs to consider the presence of other obstacles, and the interactions with them will affect the resulting behaviour. In particular, the agent's velocity is affected by possible collisions (e.g. sudden stop or direction change to avoid an obstacle). Hence, the expected causal links in this scenario are the following ones:
\vspace{-5pt}
\begin{enumerate}[label=(\alph*)]
    \item $d_g$ depends inversely on $v$;
    \item $v$ is a direct function of the $d_g$, but it is also affected by the collision $risk$;
    \item $risk$ depends on the velocity, as explained below.
\end{enumerate}
\textbf{Obstacle detection and risk evaluation:} in the literature, there are several strategies for identifying obstacles and evaluating the risk of collision with them. In order to model a numerical $risk$ value as a function of the agent's interactions, we implemented a popular strategy named Velocity Obstacles~(VO)~\cite{Fiorini1998}. The VO technique identifies an unsafe sub-set of velocities for the selected agent that would lead to a collision with a moving or static obstacle, assuming the latter maintains a constant velocity.

\begin{figure}[t]\centering
\includegraphics[width=0.66\columnwidth]{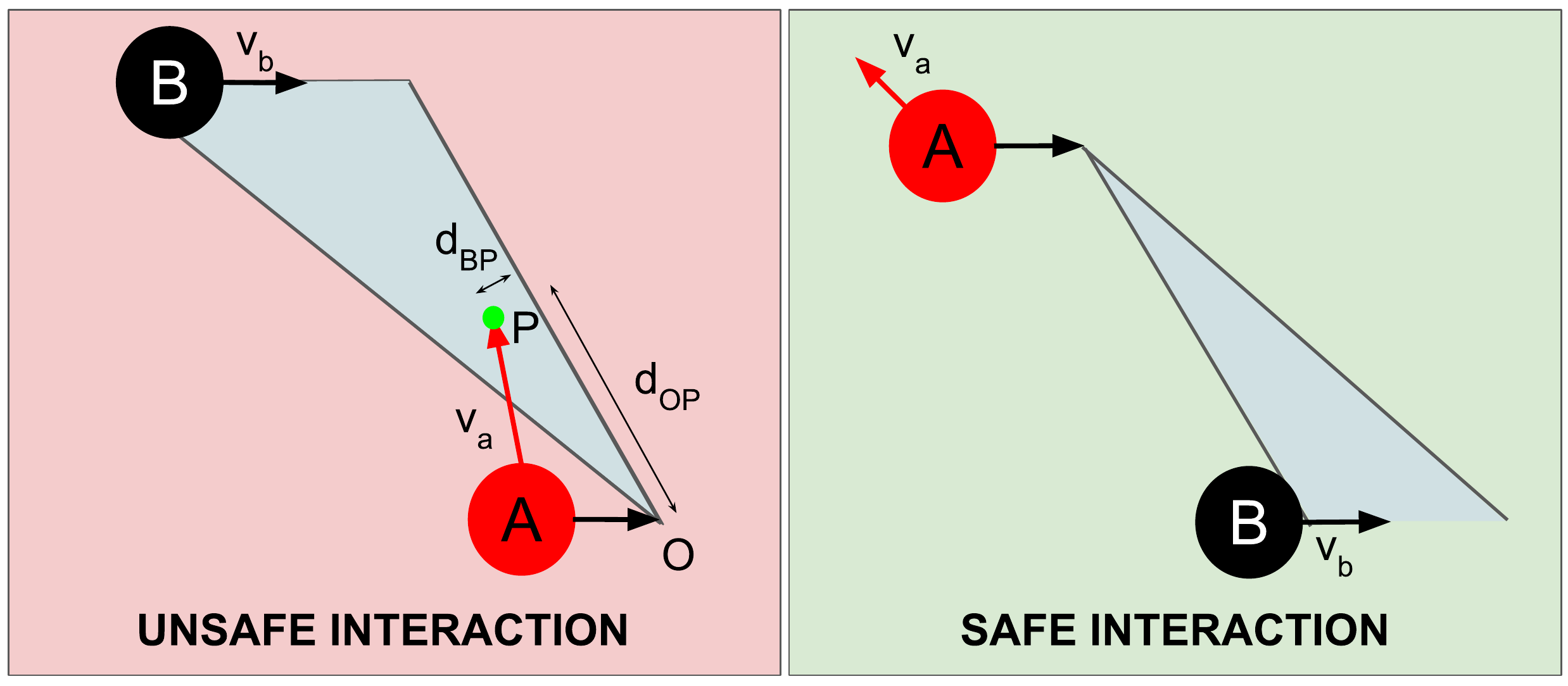}
\caption{\small{Velocity Obstacle (VO) technique. A Collision Cone (CC) is built from the selected agent A to the enlarged encumbrance of the obstacle B. Then, the CC is translated by $v_b$ to identify the VO, which partitions the velocity space of A into \textit{avoiding} and \textit{colliding} regions, i.e. velocities lying outside and inside the VO, respectively. (Left) an interaction leading to a collision. (Right) a collision-free interaction.}}
\vspace{\negspace}
\label{fig:approach_velocity_obstacle}
\end{figure}

The risk can then be defined as follows. At each time step, we apply the VO to the agent's closest obstacle. Such risk is a function of two parameters, both depending on the selected agent's velocity (i.e. point $P$ inside the VO; see Fig.~\ref{fig:approach_velocity_obstacle}):
\vspace{-5pt}
\begin{itemize}
    \item $d_{OP}$, the distance between the cone's origin $O$ and $P$, which is proportional to the time available for the selected agent A to avoid the collision with B;
    \item $d_{BP}$, the distance between $P$ and the closest cone's boundary, which indicates the steering effort required by A to avoid the collision with B.
\end{itemize}
\vspace{-5pt}
Consequently, the risk of collision is defined as follows: 
\begin{equation}
    risk = e^{d_{OP} + d_{BP} + v_a}.
\end{equation}
In order to avoid mostly-constant values (undetectable by the causal discovery algorithm), we introduced a third parameter $v_a$, which is the velocity of the selected agent. Therefore, the risk depends mainly on the agent's velocity, plus the VO's contributions in case of interaction with another agent.

\subsection{Causal prediction with PCMCI and GPR}
Our approach for modeling and predicting spatial interactions, shown in Fig.~\ref{fig:intro}, can be decomposed in three main steps: (\textit{i})~extract the necessary time-series of sensor data from the two previously explained scenarios; (\textit{ii})~use them for the causal discovery performed by the PCMCI algorithm; (\textit{iii})~finally, embed the causal models in a GPR-based prediction system.
More in detail, PCMCI is a causal discovery algorithm~\cite{runge_causal_2018} which consists of two main parts, both exploiting conditional independence tests (e.g. partial correlation, Gaussian processes and distance correlation) to measure the causal strength between variables. The first part is the well-known PC algorithm, which starts from a fully connected graph and outputs an initial causal model structure; the latter is then used by the second part, the MCI test, which validates the structure by estimating the test statistics values and p-values for all the links and outputs the final causal model. After that, we exploit the GPR, a nonparametric kernel-based probabilistic model~\cite{li2020data}, to build a causal GPR predictor, useful to forecast each variable by using only its parents, and not all the variables involved in the scenario, as a non-causal GPR predictor would do.

\section{Experiments} \label{sec:exp}
We evaluated our approach for causal modeling and prediction of human spatial behaviours on two challenging datasets: TH{\"O}R~\cite{thorDataset2019} and ATC Pedestrian Tracking~\cite{brvsvcic2013person}. Both contains data of people moving in indoor environments, a workshop/warehouse and a shopping center, respectively. Our strategy is first to extract the necessary time-series from the two datasets, as explained in Sec.~\ref{sec:appr}, and then use it for causal discovery. In order to prove the usefulness of the causal models, a comparison between causal and a non-causal predictions is finally shown.
We considered two different datasets in order to verify, for the first scenario in Sec.~\ref{subsec: Human-goal scenario}, that the discovered causal model holds for similar human behaviours, even when observed in different environments. The scenario in Sec.~\ref{subsec: Human-moving obs scenario}, instead, is used to demonstrate that it is possible to perform causal discovery for other types of human spatial interactions (i.e. with collision avoidance).

\subsection{Data processing}
From both datasets, we extracted the $x$-$y$ positions of each agent and derived all the necessary quantities from them (i.e. orientation $\theta$, velocity $v$, etc.).\\
\textbf{TH{\"O}R dataset:} this provides a wide variety of interactions between humans, robot, and static objects~(Fig.~\ref{fig:approach_scenarios}, left). Helmets and infrared cameras were used to track the motion of the agents at~$100~Hz$. We used this dataset to analyse both scenarios (Sec.~\ref{subsec: Human-goal scenario} - \ref{subsec: Human-moving obs scenario}). 
Moreover, to reduce the computational cost of causal discovery on this dataset, due to the high sampling rate, we subsampled the dataset using an entropy-based adaptive-sampling strategy~\cite{Aldana-Bobadilla2015}, with an additional variable size windowing approach to reduce
the number of samples.\\
\textbf{ATC pedestrian tracking dataset:} in this case, the data was collected in the large atrium of a shopping mall (much bigger than TH{\"O}R's environment). Several 3D~range sensors were used to track people at $30~Hz$. Due to its large area and crowd, this dataset was not suitable for the collision-enhanced scenario in Sec.~\ref{subsec: Human-moving obs scenario}. Indeed, the large distance between humans and goals made the VO and the risk analysis difficult to estimate. Therefore, we used this dataset only for the scenario in Sec.~\ref{subsec: Human-goal scenario}, assuming that the interactions and collision avoidance between humans could be captured by the model's noise variance.

\subsection{Results}
\begin{figure*}[t]
\centering 
\subfloat{
    \includegraphics[trim={2.5cm 1cm 2.5cm 1.3cm}, clip, width=0.33\textwidth]{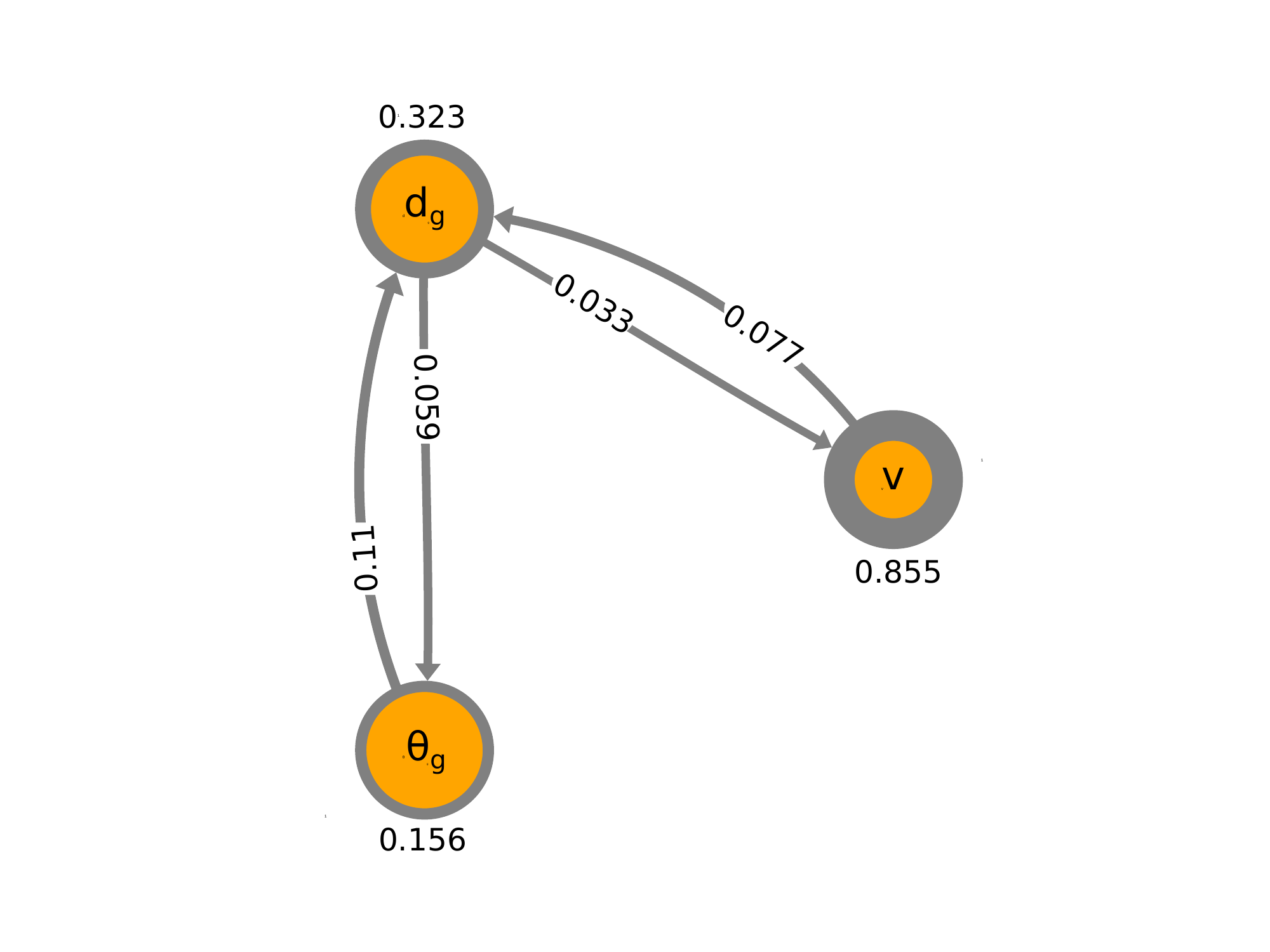}}
\subfloat{
    \includegraphics[trim={2.5cm 1cm 2.5cm 1.3cm}, clip, width=0.33\textwidth]{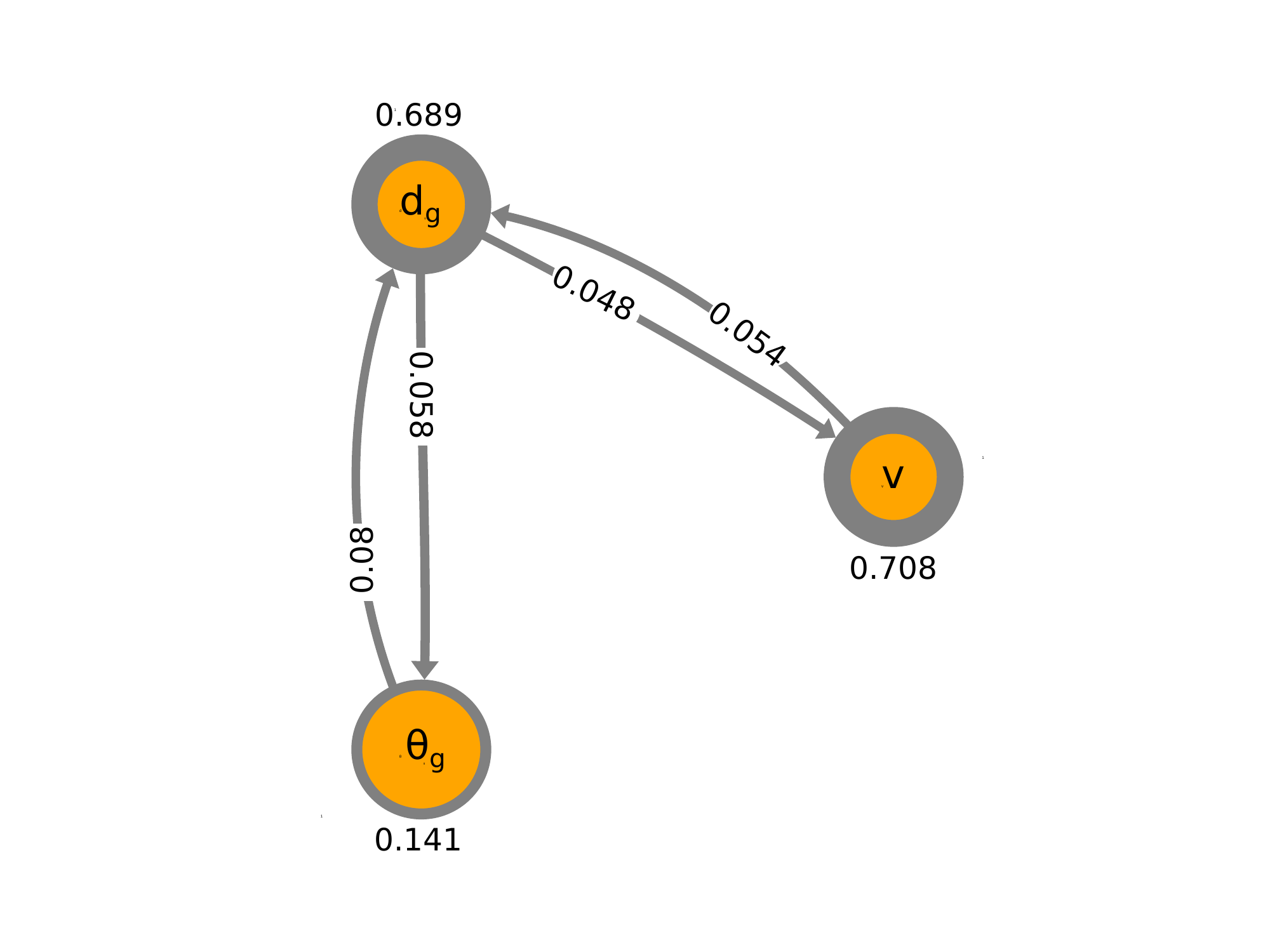}}
\subfloat{ 
    \includegraphics[trim={2.5cm 1cm 2.5cm 1.3cm}, clip, width=0.33\textwidth]{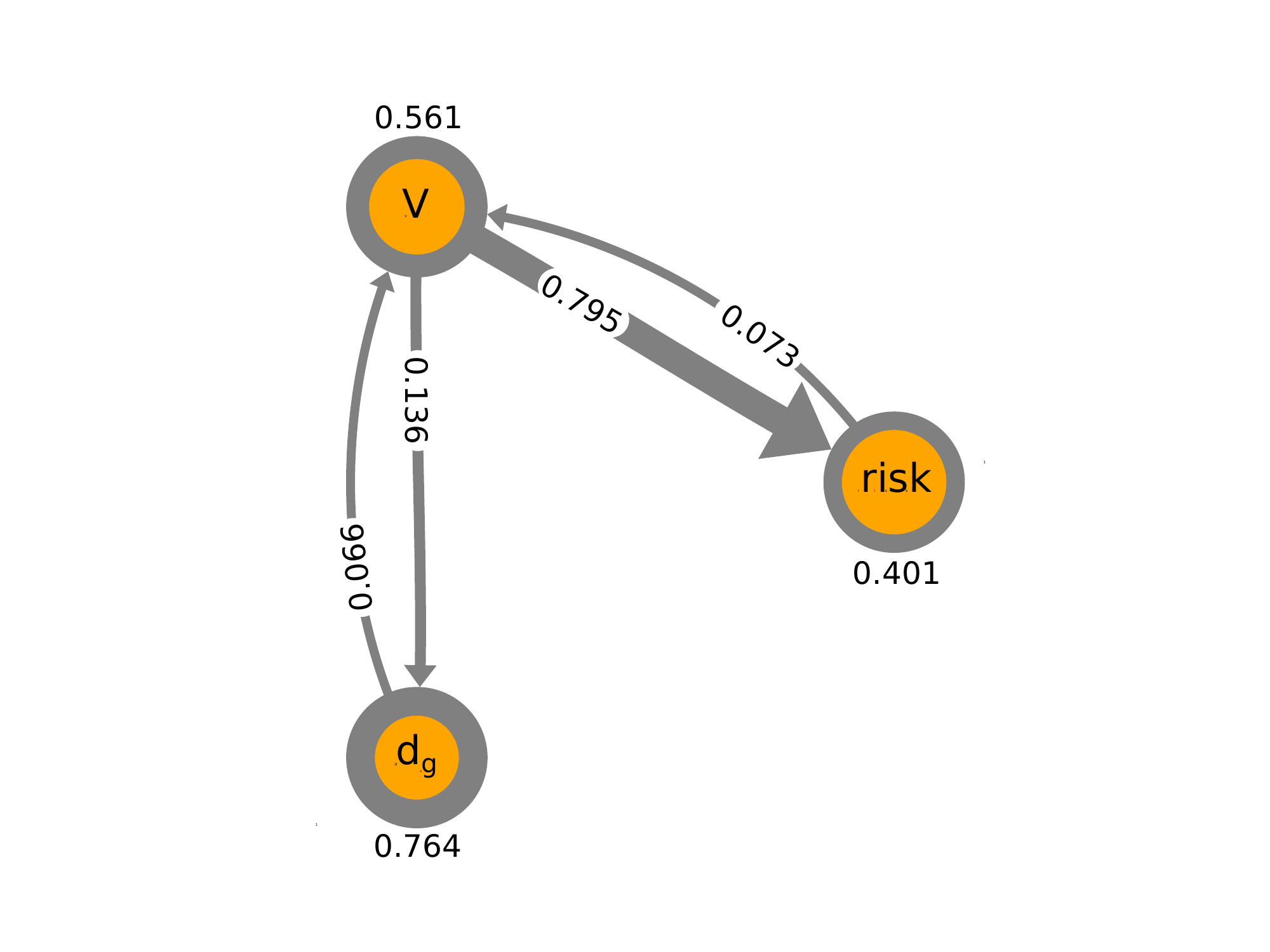}}
\vspace{-7pt}
\caption{\small{Causal models of: human-goal scenario with (left) TH{\"O}R and (centre) ATC datasets, human-moving obstacles scenario with (right) TH{\"O}R dataset. The thickness of the arrows and of the nodes’ border represents, respectively, the strength of the cross and auto-causal dependency, specified by the number on each node/link (the stronger the dependency, the thicker the line). All the relations correspond to a 1-step lag time.}}
\vspace{\negspace}
\label{fig:exp_causal_model}
\end{figure*}

We report the causal models discovered by PCMCI for the two scenarios. The latter were obtained using the same conditional independence test based on Gaussian Process regression and Distance Correlation (GPDC)~\cite{runge_detecting_2019}. We used also a 1-step lag time, that is, variables at time $t$ could only be affected by those at time $t-1$.
%
The resulting causal models are shown in Fig.~\ref{fig:exp_causal_model}, where the thickness of the arrows and of the nodes’ border represents, respectively, the strength of the cross and auto-causal dependency, specified by the number on each node/link (the stronger the dependency, the thicker the line).

In particular, Fig.~\ref{fig:exp_causal_model} shows the causal models of the human-goal~(left and centre) and the human-moving obstacles scenarios~(right). All the three graphs agree with the expected models discussed in Sec.~\ref{subsec: Human-goal scenario} and Sec.~\ref{subsec: Human-moving obs scenario}. This confirms that the same causal structure in the first two graphs generalises to similar human behaviours in different datasets, although causal strengths vary between them due to different sampling frequencies and noise levels. The third graph proves that it is possible to get different causal models for different human behaviours.

\begin{figure*}[t]
\centering 
\subfloat{
    \label{fig:exp_pred_NMAE_theta}
    \includegraphics[trim={0.8cm 0.5cm 2cm 1cm}, clip, width=0.44\textwidth]{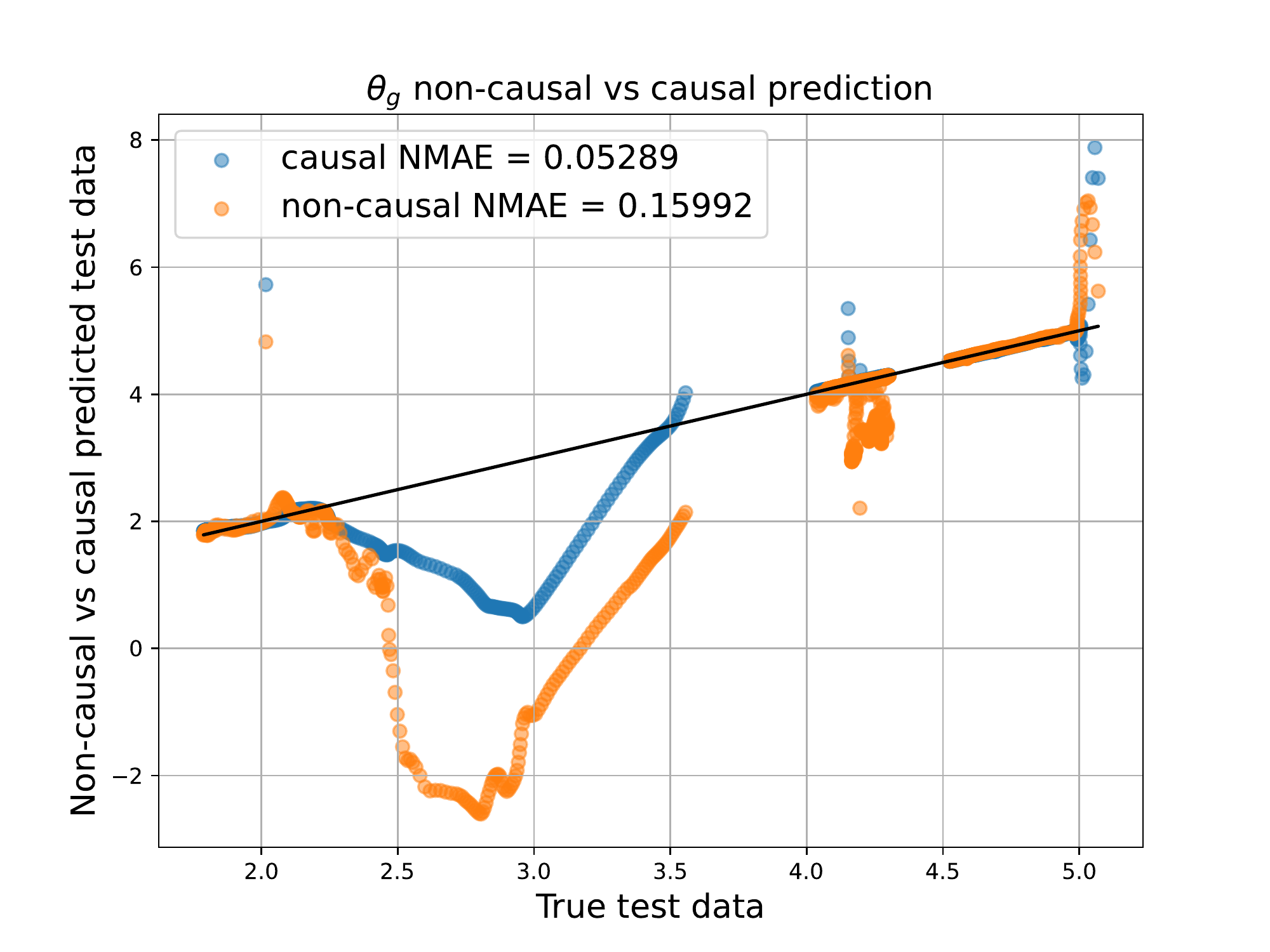}}
\subfloat{
    \label{fig:exp_pred_NMAE_dg}
    \includegraphics[trim={0.8cm 0.5cm 2cm 1cm}, clip, width=0.44\textwidth]{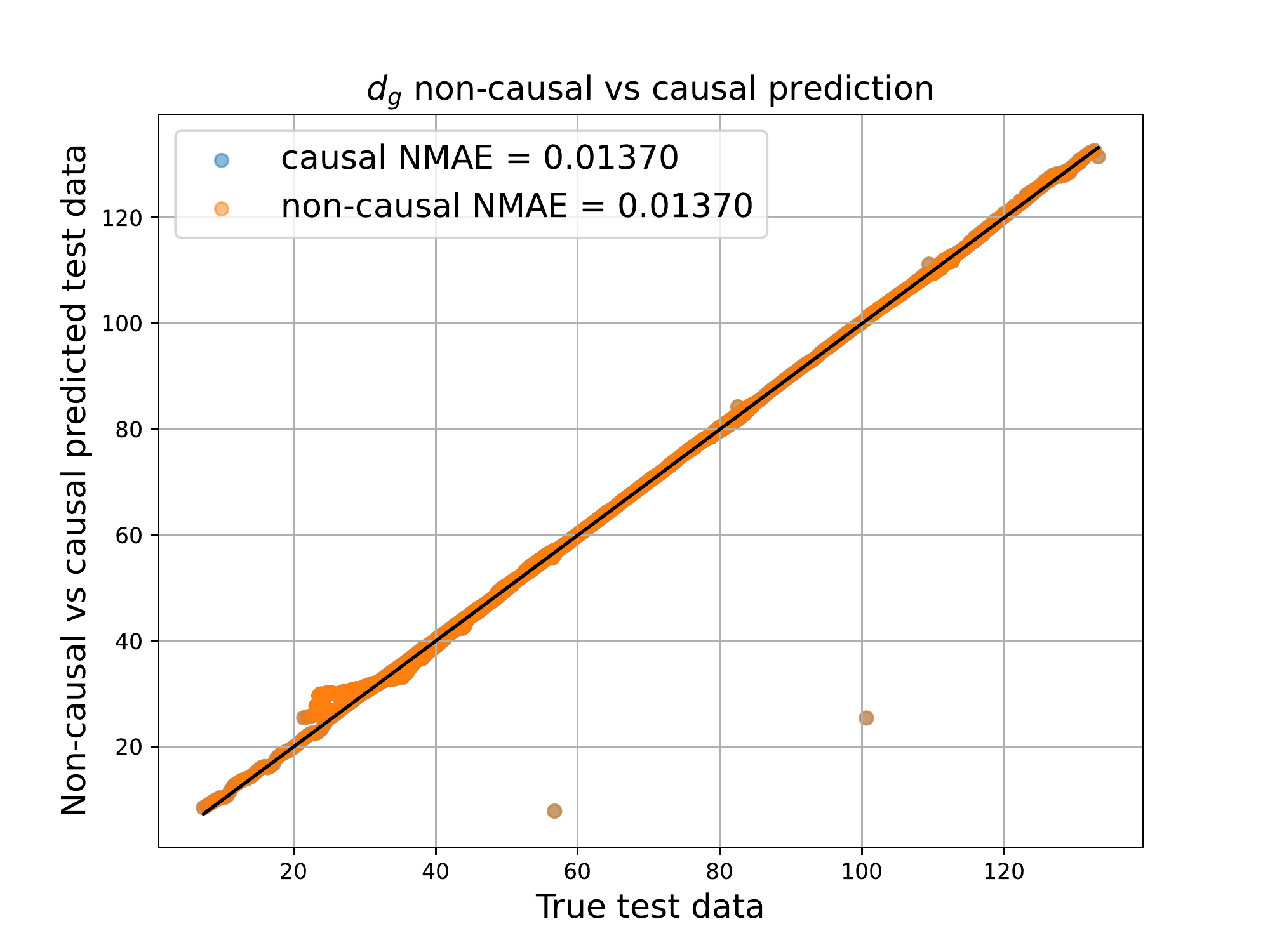}}\\
\subfloat{
    \label{fig:exp_pred_NMAE_v}
    \includegraphics[trim={0.8cm 0.5cm 2cm 1cm}, clip, width=0.44\textwidth]{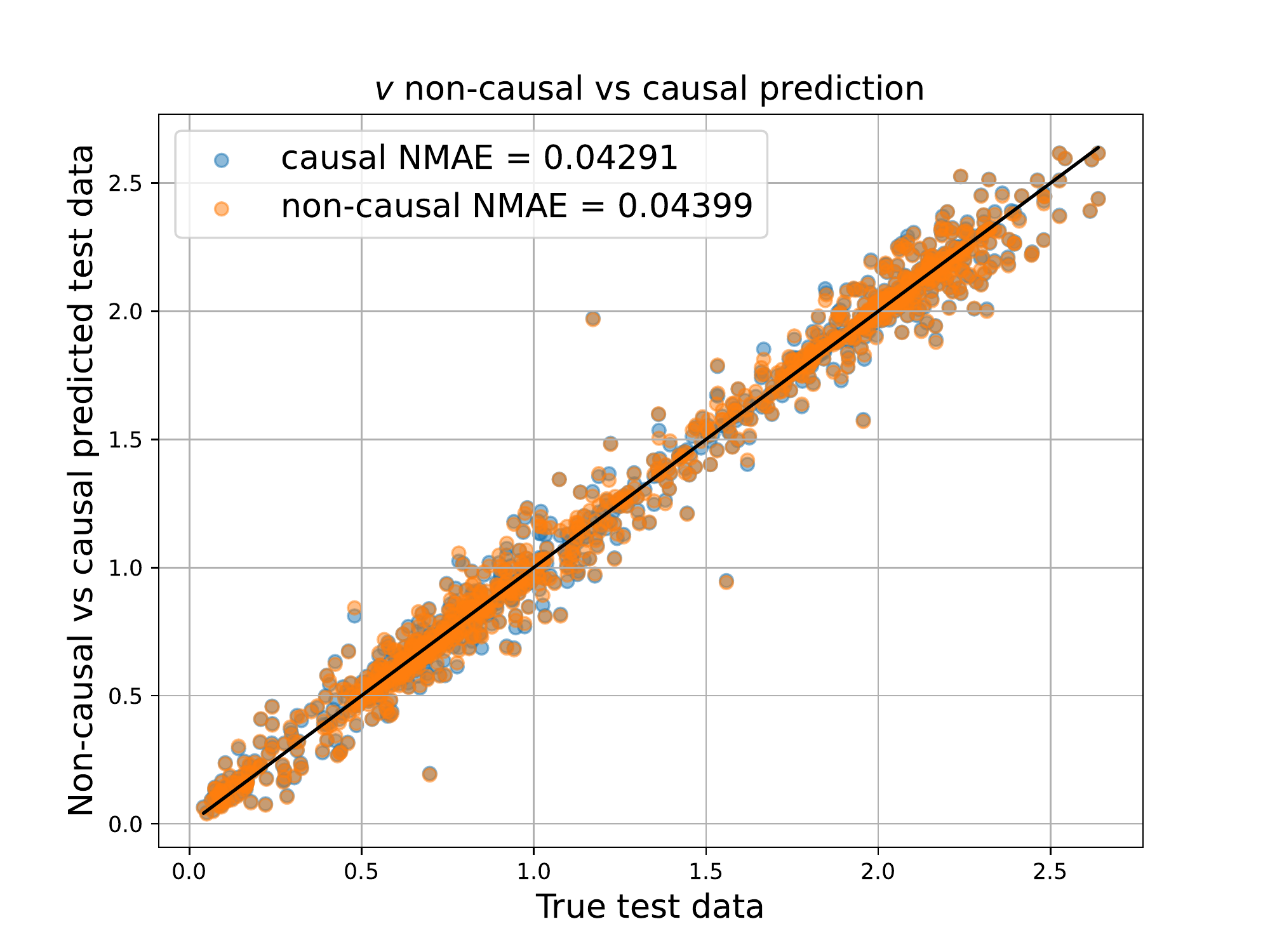}}
\subfloat{
    \label{fig:exp_pred_NMAE_bar}
    \includegraphics[trim={0cm 0cm 0cm 0cm}, clip, width=0.44\textwidth]{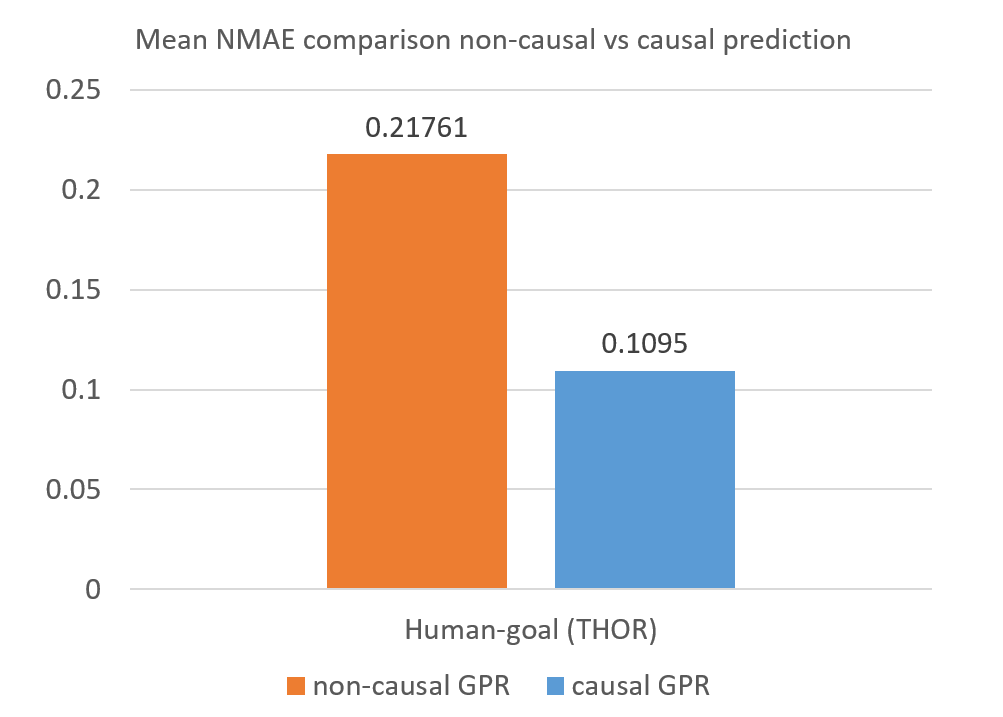}}
\vspace{-7pt}
\caption{\small{Comparison between non-causal and causal GPR prediction and NMAE in the human-goal scenario of the TH{\"O}R dataset for the spatial interaction variables $\theta_g$~(top-left), $d_g$~(top-right), and $v$~(bottom-left). A bar chart (bottom-right) summarises the comparison using the mean NMAE over all three variables.}}
\vspace{\negspace}
\label{fig:exp_pred}
\end{figure*}

The obtained causal models are then exploited for the prediction of the spatial interaction variables in both scenarios. For example, in case of the human-goal scenario using the TH{\"O}R dataset, the prediction of $\theta_g$ was done using only its parents $\theta_g$ and $d_g$ in the respective causal model (Fig.~\ref{fig:exp_causal_model}, left). To evaluate the advantage of using our models, we benchmarked the prediction performance with a causally-informed GPR estimator against a non-causal one, where the latter considers all the variables possibly influencing each other. Fig.~\ref{fig:exp_pred} shows the comparison between causal and non-causal prediction for the human-goal scenario with the TH{\"O}R dataset, using as evaluation metric the \textit{Normalised Mean Absolute Error}~(NMAE), which is not too sensitive to possible outliers. This is defined as follows: 
\begin{equation}
NMAE(y, \hat{y}) = \frac{\sum_{i=1}^{n}\frac{\lvert y_i - \hat{y_i}\rvert}{n}}{\frac{1}{n}\sum_{i=1}^{n} y_i}
\end{equation}
where $y$ and $\hat{y}$ are, respectively, the actual and the predicted values. Fig.~\ref{fig:exp_pred}~shows that our causal model helps to predict the variables $\theta_g$~(top-left) and $v$~(bottom-left) more accurately compared to the non-causal case. Indeed, the NMAE of the causal predictor is lower than the non-causal one. Instead, for the variable $d_g$~(top-right) the predictors set corresponds to the full set of predictors in both causal and non-causal approaches, leading to the same NMAE results. Finally, Fig.~\ref{fig:exp_pred}~(bottom-right) shows a bar chart summarising the NMAE comparison over all the three variables, showing that the causal model's knowledge helps the GPR to predict the system more accurately.
In conclusion, Table~\ref{tab:exp_NRMSE} reports the above-explained analysis for all the considered scenarios, highlighting that the causal GPR approach improves always the prediction accuracy compared to the non-causal one. Note that, for the human-goal scenario, the mean NMAE in the ATC dataset is bigger than in TH{\"O}R, which is probably due to the different time-series lengths in the two datasets.

\begin{table*}[t]
\centering
\begin{tabular}{L{2cm}|C{2cm}|C{2cm}|C{4cm}|}
& \multicolumn{2}{c|}{Human-goal} & Human-moving obs \\ \cline{2-4} 
& TH{\"O}R & ATC & TH{\"O}R \\ \hline
Non-causal & 0.21761 & 1.61692 & 0.37849 \\ \hline
Causal     & \textbf{0.1095}  & \textbf{1.54552} & \textbf{0.36453} \\ \hline
\end{tabular}
\vspace{2mm}
\caption{\small{Mean NMAE of causal and non-causal predictions over the involved variables for both scenarios and datasets.}}
\vspace{-1cm}
\label{tab:exp_NRMSE}
\end{table*}

\section{Conclusion} \label{sec:conc}
In this work, we proposed a causal discovery approach to model and predict Human Spatial Interactions. We used two public datasets (TH{\"O}R and ATC) to extract time-series of human motion behaviours in two possible scenarios for causal analysis. We show that the discovery algorithm can capture the expected causal relations from the datasets. We used the obtained causal models to predict the values of some key spatial interaction variables, and benchmarked them against the results of a non-causal prediction approach. The comparison highlights the contribution and the advantage of integrating such causal models in the prediction framework.
Future work will be devoted to automatically learn the most important features for modelling human-human and human-robot spatial interactions, focusing in particular on on-board robot sensor data. We will also perform causal reasoning on both observational and interventional data, exploiting the influence that the robot's presence can have on nearby people, with a special interest for applications in industrial and intralogistics settings.

\bibliographystyle{splncs04}
\bibliography{references_new}

\end{document}